\documentclass{article}

\PassOptionsToPackage{numbers, compress}{natbib}
\usepackage[final]{paperstyle}

\usepackage[utf8]{inputenc}
\usepackage[T1]{fontenc}
 
\usepackage{hyperref}
\usepackage{url}
\usepackage{booktabs}
\usepackage{amsfonts}
\usepackage{amsmath}
\usepackage{amssymb}
\usepackage{nicefrac}
\usepackage{microtype}
\usepackage{xcolor}
\usepackage{graphicx}
\usepackage{multirow}
\usepackage{placeins}

\graphicspath{{figures/}}


\title{Co-folding model guided by structural proteomics}

\author{
  Alon Shtrikman$^{1*}$ \quad
  Sagie Brodsky$^{1*}$ \quad
  Michal Ran Shchory$^{1*}$ \quad
  Nitzan Simchi$^{1*}$ \\[0.15cm]
  Yaron Ben Shoshan-Galeczki$^{1}$ \quad
  Eran Seger$^{1}$ \quad
  Kirill Pevzner$^{1}$ \\[0.2cm]
  $^1$Protai Bio \\[0.15cm]
  \small{$^*$These authors contributed equally \quad
         Correspondence: \texttt{Kirill@protai.bio}}
}

\newcommand{\figscale}{0.8}
\makeatletter\renewcommand{\@toptitlebar}{\vskip 0.1in}\renewcommand{\@bottomtitlebar}{\vskip 0.12in}\makeatother
\AtBeginDocument{\newgeometry{textheight=9.4in,textwidth=6.25in,top=0.85in,headheight=12pt,headsep=18pt,footskip=24pt}}
\renewenvironment{abstract}{\vskip 0.075in\centerline{\large\bf Abstract}\vspace{0.5ex}\begin{list}{}{\setlength{\leftmargin}{0pt}\setlength{\rightmargin}{0pt}}\item[]\relax}{\par\end{list}\vskip 1ex}
\let\rawgraphics\includegraphics\renewcommand{\includegraphics}[2][]{\scalebox{\figscale}{\rawgraphics[#1]{#2}}}
\begin{document}
\maketitle

\begin{abstract}
Proximity-inducing drugs, including PROTACs, represent a promising therapeutic modality. Their activity is driven by the geometry of the assembled ternary complex, which determines whether the target is ubiquitinated. Co-folding models perform poorly in these cases due to scarcity of training data and the dynamic nature of such complexes. Structural proteomics can supply the missing geometric information and steer the model towards an accurate prediction. However, current models accept only positive signals, while a differential structural mass-spectrometry experiment also reports negative signals. We introduce AIMS-Fold, an inference-time guidance layer built on Boltz-2 that requires no retraining and accepts repulsive constraints alongside attractive ones. We tested AIMS-Fold on 28 PROTAC, glue and antibody complexes that Boltz-2 fails to predict. From only five generated models, guidance recovers the correct conformation in 64\% of cases. The gap does not close with scale: across 100 models and 20 seeds, unguided Boltz-2 reaches 14\% while AIMS-Fold reaches 86\%. Repulsive constraints alone recover the correct conformation on some systems, occasionally exceeding the matched attractive constraints. Taken together, these results demonstrate the strength of utilizing the full structural proteomics signal, positive and negative, to steer generative models toward the correct conformation, addressing a central problem in rational degrader design.
\end{abstract}
\begin{center}\begin{minipage}{\linewidth}
{\centering
{\renewcommand{\figscale}{1}\includegraphics[width=\linewidth,height=0.43\textheight,keepaspectratio]{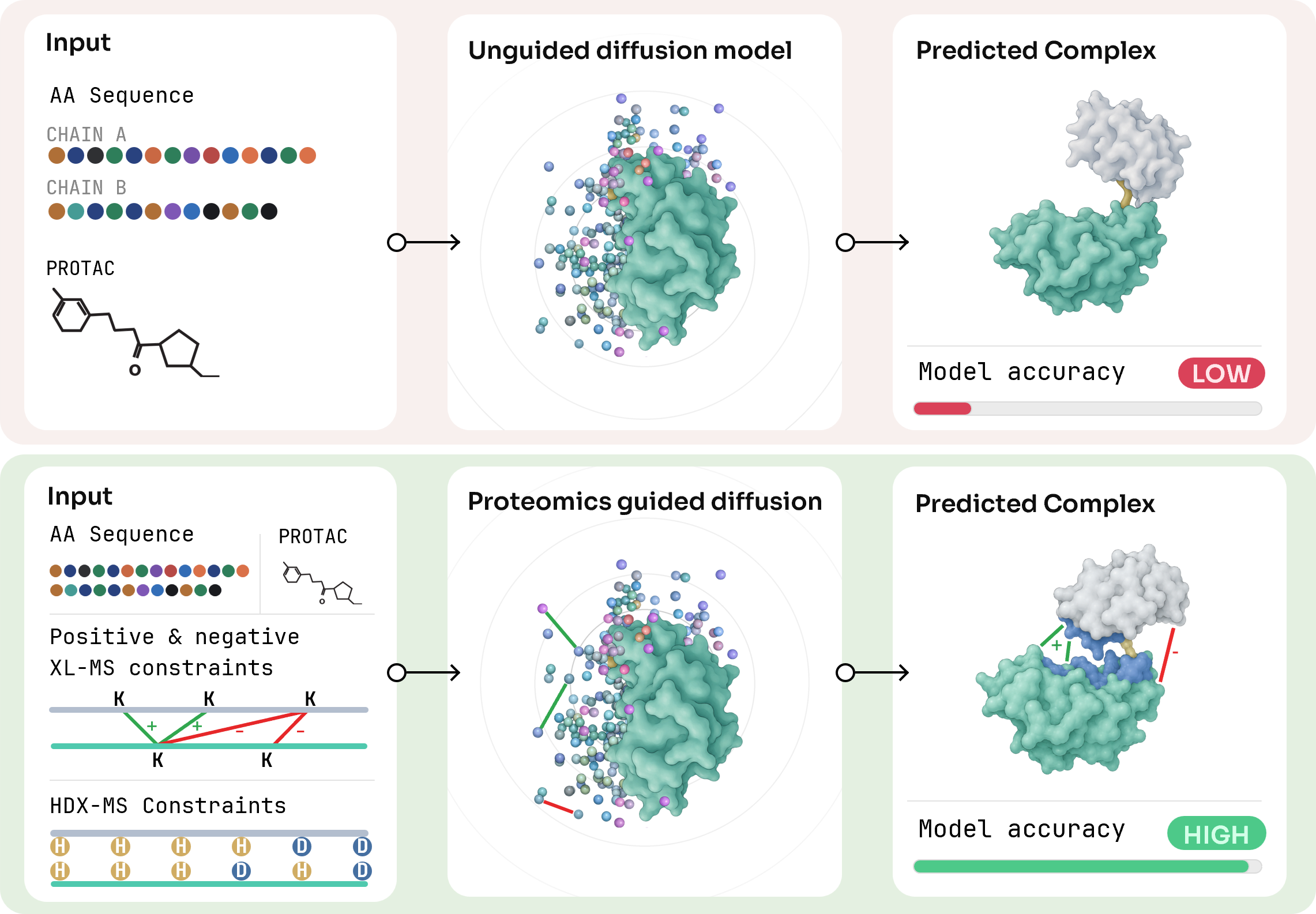}}\par}\vspace{1ex}
\noindent{\small\textbf{Graphical Abstract.} Schematic illustration of AIMS-Fold. Top, unguided Boltz-2 receives only the protein sequences and the PROTAC and can converge on an incorrect ternary complex. Bottom, AIMS-Fold is an inference-time guidance layer on Boltz-2 that also receives attractive and repulsive XL-MS constraints and HDX-MS constraints. It uses them to steer the diffusion trajectory toward the correct conformation during generation, rather than filtering models afterwards, and needs no retraining.}
\end{minipage}\end{center}
\section{Introduction}

Proximity-inducing drugs, including proteolysis-targeting chimeras (PROTACs) and molecular glues, represent a promising class of therapeutic modalities \cite{Bekes2022,Schreiber2021}. Unlike classical small molecule inhibitors whose efficacy is dictated by the binary binding affinity to a single target, the activity of PROTACs and glues is also driven by the 3D geometry of the assembled ternary complex \cite{Ward2023-dk,Wurz2023-kj,Rui2023-zb,Hughes2017-pq}. Target lysines need to be accessible and in proximity to the ligase machinery for ubiquitination to occur \cite{Nowak2018,Chamberlain2019}. Thus, ternary complex formation does not necessarily lead to degradation. For rational drug design, where an accurate prediction of the protein complex is critical, the question then becomes which conformation is formed and not whether one forms \cite{Ignatov2023-gd}.

Recently, sequence-to-structure deep learning models, such as AlphaFold3 \cite{Abramson2024} and Boltz-2 \cite{Passaro2025}, have impacted structural biology by providing highly accurate all-atom proteome-wide structure predictions. Despite these breakthroughs, the task of predicting interfaces of complexes formed by PROTACs and molecular glues has proven hard \cite{Xue2025,Pereira2025-va,Liao2025}. This is due to the scarcity of available training data for the models \cite{Xue2025,Liao2025}, the small interfaces which leave the relative orientation of the two proteins underdetermined \cite{Pereira2025-va}, and, for PROTACs, the dynamic nature of the complexes \cite{Rovers2024}. The same challenge holds for antibody-antigen complexes, where the sequence does not indicate which surface region of the antigen is bound, and prediction accuracy remains among the weakest of any complex type \cite{Hitawala2025}.

Unlike drug-pocket optimization, where sub-angstrom accuracy is required to maximize localized binding affinity and inhibition, PROTACs pose a distinct spatial challenge. Successful PROTAC-induced degradation hinges on global structural arrangement: Bringing target lysines into the effective zone of the E3 ligase, an event operating over a range of tens of \AA ngstr\"oms \cite{Fischer2011,Bai2022}. The modeling goal therefore shifts from fine atomic docking to macro-level spatial orientation \cite{Pereira2025-va}. Similarly, in antibody design, identifying which epitope is engaged is a question of surface localization rather than atomic precision.

Where the model fails to predict the correct structure, experimental data can supply what is missing, and structural proteomics is well suited to this. XL-MS (Cross-Linking Mass Spectrometry) reports spatial proximity between two specific residues \cite{Lee2023}. While the collected data is sparse, cross-links spanning the two proteins are paired measurements between specific residues, providing clues for the relative orientation of the chains. HDX-MS (Hydrogen-Deuterium Exchange Mass Spectrometry) reports which regions become protected or remain solvent exposed upon complex formation \cite{Masson2019}. While denser than cross-linking data, HDX-MS data is unpaired, and at peptide resolution, providing information regarding where the interface is located without reporting what it touches.

XL-MS and HDX-MS can complement each other since the first narrows the orientation search space, while the latter locates the interface. In addition, using both modalities can be beneficial when data generated by one alone is not enough to steer the model to the right conformation. Both techniques can be used to compare across different conditions, e.g. comparing an active compound to an inactive one or comparing ternary complex to the apo state, resulting in a differential signal between conformations. Notably, both methods measure the protein in solution, representing a dynamic ensemble rather than a single crystallized or frozen state \cite{Lee2023,Masson2019}.

Structural MS data is commonly used for post-hoc pose ranking and scoring \cite{Dominguez2003,Tran2025-hb,Roberts2013-ot,Wang2025-lu}, which can only select among the structures the sampler produced. Recent generative models accept cross-links as input constraints \cite{Stahl2023,Stahl2024}, and HDX data can be supplied indirectly through pocket constraints \cite{Chai2024}. Inference-time guidance, which requires no retraining, has also been used to steer structure prediction with cryo-EM density \cite{Raghu2025}. The transition from post-hoc scoring to data integration throughout the generative step represents significant progress in the field. However, these frameworks still only accept positive signals. Both XL-MS and HDX-MS also contain negative signals that are usually overlooked: cross-links showing weaker signal in potent versus inactive PROTACs, and regions that show no differential deuterium uptake despite good peptide coverage. Given that a single experiment does not always yield enough constraints, these should not be discarded. Neither, however, can currently be utilized by existing co-folding models, which accept only attractive constraints and cannot represent repulsive ones that push residues apart.

We introduce AIMS-Fold, an inference-time guidance layer built on Boltz-2 that requires no retraining. On top of the attractive cross-links that are already supported, our model accepts repulsive XL-MS and both attractive and repulsive HDX-MS signals, with the latter entering through the model's pocket constraints. Differential HDX-MS identifies the buried region but does not always give the information about the chain that buries it, whereas a Boltz-2 pocket constraint requires a single named binder chain. We therefore extended the constraint to accept a set of candidate partner chains, satisfied when any of them covers the region. AIMS-Fold is, to our knowledge, the first co-folding model to accept repulsive signals, which are routinely produced in drug discovery campaigns.
Using AIMS-Fold, we show that guidance recovers the correct conformation where the unguided model fails, that repulsive constraints alone can suffice and that more samples do not close the gap.

\section{Results}

\subsection{Guided structure generation succeeds where Boltz-2 fails}

We collected 28 complexes with a reference structure for which Boltz-2 fails to reproduce the correct pose, along with a novel complex, KAT6A-CRBN, which has no solved structure. We ran full constraint ablations on five of them, chosen because three have measured structural MS data (PD-1-Nivolumab, PTPN2-CRBN, and the in-house KAT6A-CRBN) and two were solved in two conformations with different PROTACs (BRD4-CRBN and WDR5-DCAF1), where Boltz-2 locks onto one even when given the other's compound. The other 24 were run with the full constraint set only (Methods). For each condition we report the model with the highest percentage of fulfilled constraints out of five generated. Metrics and constraint selection per case are provided in the Supplementary materials.

\begin{figure}[ht]
    \centering
    \includegraphics[width=\linewidth]{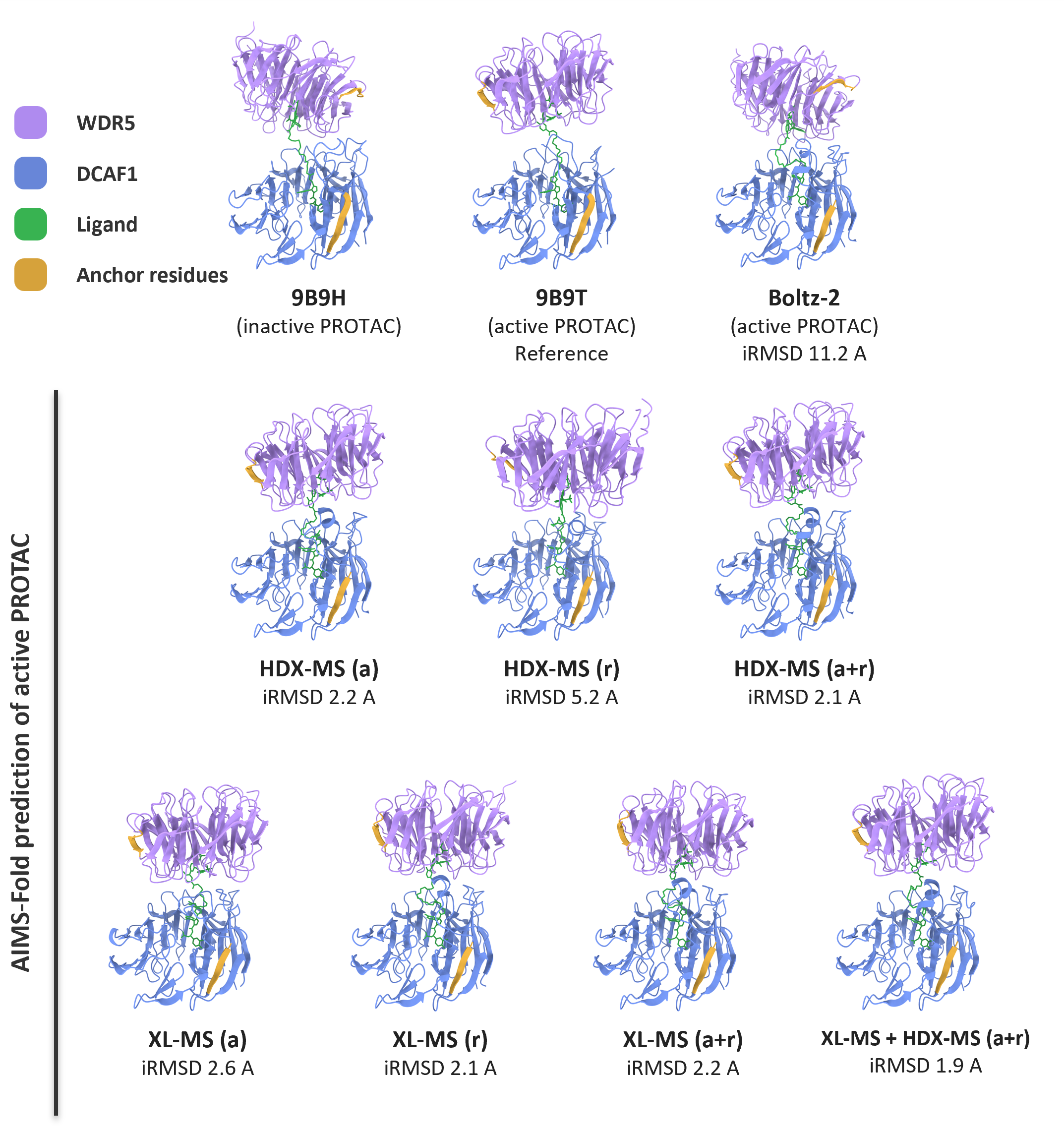}
    \caption{WDR5-PROTAC-DCAF1 ternary complex predictions. All structures are superposed on DCAF1 and shown in a common orientation. WDR5 is coloured purple, DCAF1 blue, and the PROTAC in green. Gold stretches mark a reference segment on each chain, included for easy comparison across panels. The top row shows the two crystal structures of the complex, 9B9H and 9B9T, which place WDR5 in different orientations relative to DCAF1, followed by the unguided Boltz-2 prediction for the compound of 9B9T. The prediction resembles 9B9H. 9B9T is the reference against which all predictions are scored. The panels below are AIMS-Fold predictions of the 9B9T complex under different constraint sets: HDX-MS and XL-MS, each applied as attractive only (a), repulsive only (r), or both together (a+r), and the two modalities combined in their attractive and repulsive forms. Interface RMSD to 9B9T is given below each model.}
    \label{fig:wdr5}
\end{figure}

The two solved structures of WDR5-DCAF1, 9B9H and 9B9T, were obtained with two different PROTACs and show distinctly different orientations of WDR5 relative to DCAF1 \cite{Mabanglo2024}. Boltz-2 fails to capture this PROTAC-dependent shift, favoring an orientation that resembles 9B9H even when given the compound of 9B9T (DockQ 0.06, iRMSD 11.2 \AA, Table~\ref{tab:aimsfold_benchmark_constraints} and Figure~\ref{fig:wdr5}). Attractive XL-MS constraints recover the orientation (0.48, 2.6 \AA), and the repulsive XL-MS constraints derived from links that weaken in the 9B9T orientation perform comparably on their own (0.50, 2.1 \AA). Combining the XL-MS with HDX-MS constraints further improves the prediction for this system (0.53, 1.9 \AA; Figure~\ref{fig:wdr5}).

PTPN2-CRBN has published HDX-MS data \cite{Hao2024}, which we used in AIMS-Fold. Neither the attractive nor the repulsive constraints are sufficient on their own. The attractive constraints barely cross the acceptable threshold (DockQ 0.26, iRMSD 3.5 \AA, Table~\ref{tab:aimsfold_benchmark_constraints}, and Figure~\ref{fig:ptpn2}) and the repulsive constraints alone leave the prediction where the unguided model placed it (0.02, 15.6 \AA{} versus 0.03, 13.5 \AA). Supplying both together recovers the orientation (0.43, 2.6 \AA), indicating that the combination of attractive and repulsive experimental constraints is beneficial for the prediction of the correct pose.

\begin{table}[htbp]
\caption{Quantitative evaluation of AIMS-Fold and Boltz-2 (Unguided) across cases. Constraints are attractive (a), repulsive (r), or both (a+r), derived from HDX-MS, XL-MS, or their combination. For each case we report DockQ, iRMSD and lRMSD of the best-of-5 prediction, and the percentage of satisfied constraints by constraint type. To identify the best-of-5 prediction, models were ranked by the highest percentage of fulfilled experimental constraints, while ties were broken by DockQ, and where no reference structure was available (KAT6A) we defaulted to the first generated model among the tied candidates. The average percentage of satisfied constraints across all 5 samples is indicated together with the standard deviation, as is the average DockQ. Satisfaction is computed against the union of all constraints of a given case. DockQ quality follows CAPRI criteria. Bold values highlight a substantial improvement over the baseline model. Dashes indicate constraint types unavailable for that case.}
\label{tab:aimsfold_benchmark_constraints}
\centering
\resizebox{\textwidth}{!}{
\begin{tabular}{ll cc cc cc cc}
\toprule
& & \multicolumn{2}{c}{\textbf{DockQ} $\uparrow$} & \multicolumn{2}{c}{\textbf{Top-1 RMSD (\AA)} $\downarrow$} & \multicolumn{2}{c}{\textbf{HDX-MS union (\%)}} & \multicolumn{2}{c}{\textbf{XL-MS union (\%)}} \\
\cmidrule(lr){3-4} \cmidrule(lr){5-6} \cmidrule(lr){7-8} \cmidrule(lr){9-10}
\textbf{Case} & \textbf{Model} & top-1 & avg$\pm$std & iRMSD & lRMSD & top-1 & avg$\pm$std & top-1 & avg$\pm$std \\
\midrule
BRD4-CRBN & Unguided & 0.06 (Incorrect) & $0.06 \pm 0.00$ & 9.4 & 20.0 & 73 & $31 \pm 21.7$ & 25 & $5 \pm 10.0$ \\
 & HDX-MS (a) & \textbf{0.42 (Acceptable)} & $0.36 \pm 0.05$ & \textbf{1.7} & \textbf{4.0} & 80 & $80 \pm 0.0$ & 100 & $100 \pm 0.0$ \\
 & HDX-MS (r) & \textbf{0.43 (Acceptable)} & $0.41 \pm 0.06$ & \textbf{1.7} & \textbf{3.2} & 93 & $89 \pm 3.3$ & 100 & $100 \pm 0.0$ \\
 & HDX-MS (a+r) & \textbf{0.41 (Acceptable)} & $0.40 \pm 0.03$ & \textbf{1.8} & \textbf{4.2} & 93 & $89 \pm 5.3$ & 100 & $100 \pm 0.0$ \\
 & XL-MS (a) & 0.21 (Incorrect) & $0.19 \pm 0.05$ & 3.7 & 10.6 & 80 & $71 \pm 5.3$ & 100 & $100 \pm 0.0$ \\
 & XL-MS (r) & \textbf{0.35 (Acceptable)} & $0.20 \pm 0.12$ & \textbf{2.1} & \textbf{6.6} & 80 & $79 \pm 2.7$ & 100 & $70 \pm 36.7$ \\
 & XL-MS (a+r) & 0.16 (Incorrect) & $0.17 \pm 0.03$ & 4.4 & 11.1 & 73 & $69 \pm 3.3$ & 100 & $100 \pm 0.0$ \\
 & XL-MS (a+r) + HDX-MS (a+r) & \textbf{0.35 (Acceptable)} & $0.38 \pm 0.04$ & \textbf{1.9} & \textbf{6.0} & 93 & $88 \pm 5.0$ & 100 & $100 \pm 0.0$ \\
\midrule
WDR5-DCAF1 & Unguided & 0.06 (Incorrect) & $0.05 \pm 0.01$ & 11.2 & 29.4 & 47 & $33 \pm 8.4$ & 12 & $12 \pm 0.0$ \\
 & HDX-MS (a) & \textbf{0.47 (Acceptable)} & $0.40 \pm 0.06$ & \textbf{2.2} & \textbf{5.4} & 73 & $76 \pm 3.3$ & 100 & $82 \pm 10.0$ \\
 & HDX-MS (r) & 0.19 (Incorrect) & $0.10 \pm 0.05$ & 5.2 & 10.3 & 73 & $80 \pm 6.0$ & 88 & $62 \pm 17.7$ \\
 & HDX-MS (a+r) & \textbf{0.43 (Acceptable)} & $0.28 \pm 0.16$ & \textbf{2.1} & \textbf{4.1} & 100 & $93 \pm 6.0$ & 100 & $90 \pm 9.4$ \\
 & XL-MS (a) & \textbf{0.48 (Acceptable)} & $0.41 \pm 0.05$ & \textbf{2.6} & \textbf{4.3} & 73 & $67 \pm 4.2$ & 88 & $85 \pm 5.0$ \\
 & XL-MS (r) & \textbf{0.50 (Medium)} & $0.15 \pm 0.18$ & \textbf{2.1} & \textbf{5.4} & 73 & $65 \pm 6.5$ & 100 & $72 \pm 14.6$ \\
 & XL-MS (a+r) & \textbf{0.46 (Acceptable)} & $0.42 \pm 0.03$ & \textbf{2.2} & \textbf{6.1} & 67 & $63 \pm 9.0$ & 100 & $92 \pm 6.1$ \\
 & XL-MS (a+r) + HDX-MS (a+r) & \textbf{0.53 (Medium)} & $0.40 \pm 0.07$ & \textbf{1.9} & \textbf{2.6} & 93 & $88 \pm 6.5$ & 100 & $85 \pm 12.2$ \\
\midrule
PTPN2-CRBN & Unguided & 0.03 (Incorrect) & $0.03 \pm 0.00$ & 13.5 & 29.1 & 48 & $41 \pm 4.7$ & - & - \\
 & HDX-MS (a) & \textbf{0.26 (Acceptable)} & $0.24 \pm 0.11$ & \textbf{3.5} & \textbf{16.1} & 76 & $63 \pm 12.5$ & - & - \\
 & HDX-MS (r) & 0.02 (Incorrect) & $0.03 \pm 0.01$ & 15.6 & 32.6 & 64 & $56 \pm 6.7$ & - & - \\
 & HDX-MS (a+r) & \textbf{0.43 (Acceptable)} & $0.30 \pm 0.15$ & \textbf{2.6} & \textbf{11.4} & 76 & $74 \pm 2.0$ & - & - \\
\midrule
PD-1-Nivolumab & Unguided & 0.00 (Incorrect) & $0.00 \pm 0.00$ & 31.7 & 88.5 & 21 & $21 \pm 0.0$ & 0 & $0 \pm 0.0$ \\
 & HDX-MS (a) & \textbf{0.23 (Acceptable)} & $0.12 \pm 0.06$ & \textbf{4.4} & \textbf{12.1} & 93 & $90 \pm 3.5$ & 100 & $90 \pm 5.0$ \\
 & HDX-MS (r) & 0.00 (Incorrect) & $0.00 \pm 0.00$ & 31.7 & 88.5 & 21 & $21 \pm 0.0$ & 0 & $0 \pm 0.0$ \\
 & XL-MS (a) & \textbf{0.69 (Medium)} & $0.53 \pm 0.11$ & \textbf{1.3} & \textbf{4.6} & 86 & $76 \pm 10.7$ & 100 & $100 \pm 0.0$ \\
 & XL-MS (a) + HDX-MS (a) & \textbf{0.64 (Medium)} & $0.53 \pm 0.08$ & \textbf{1.3} & \textbf{5.2} & 93 & $90 \pm 3.5$ & 100 & $100 \pm 0.0$ \\
 & XL-MS (a) + HDX-MS (a+r) & \textbf{0.71 (Medium)} & $0.62 \pm 0.07$ & \textbf{1.2} & \textbf{5.3} & 100 & $97 \pm 3.5$ & 100 & $100 \pm 0.0$ \\
\midrule
KAT6A-CRBN & Unguided & \multicolumn{4}{c}{\multirow{6}{*}{\textit{-no available reference-}}} & 70 & $50 \pm 11.0$ & 62 & $60 \pm 5.0$ \\
 & HDX-MS (a) & \multicolumn{4}{c}{} & 70 & $70 \pm 0.0$ & 62 & $62 \pm 0.0$ \\
 & XL-MS (a) & \multicolumn{4}{c}{} & 60 & $54 \pm 4.9$ & 75 & $75 \pm 0.0$ \\
 & XL-MS (r) & \multicolumn{4}{c}{} & 70 & $62 \pm 11.7$ & 75 & $62 \pm 7.9$ \\
 & XL-MS (a+r) & \multicolumn{4}{c}{} & 60 & $56 \pm 4.9$ & 88 & $78 \pm 5.0$ \\
 & XL-MS (a+r) + HDX-MS (a) & \multicolumn{4}{c}{} & 80 & $72 \pm 4.0$ & 88 & $85 \pm 5.0$ \\
\bottomrule
\end{tabular}
}
\end{table}

\begin{figure}[t]
    \centering
    {\renewcommand{\figscale}{1}\includegraphics[width=\linewidth]{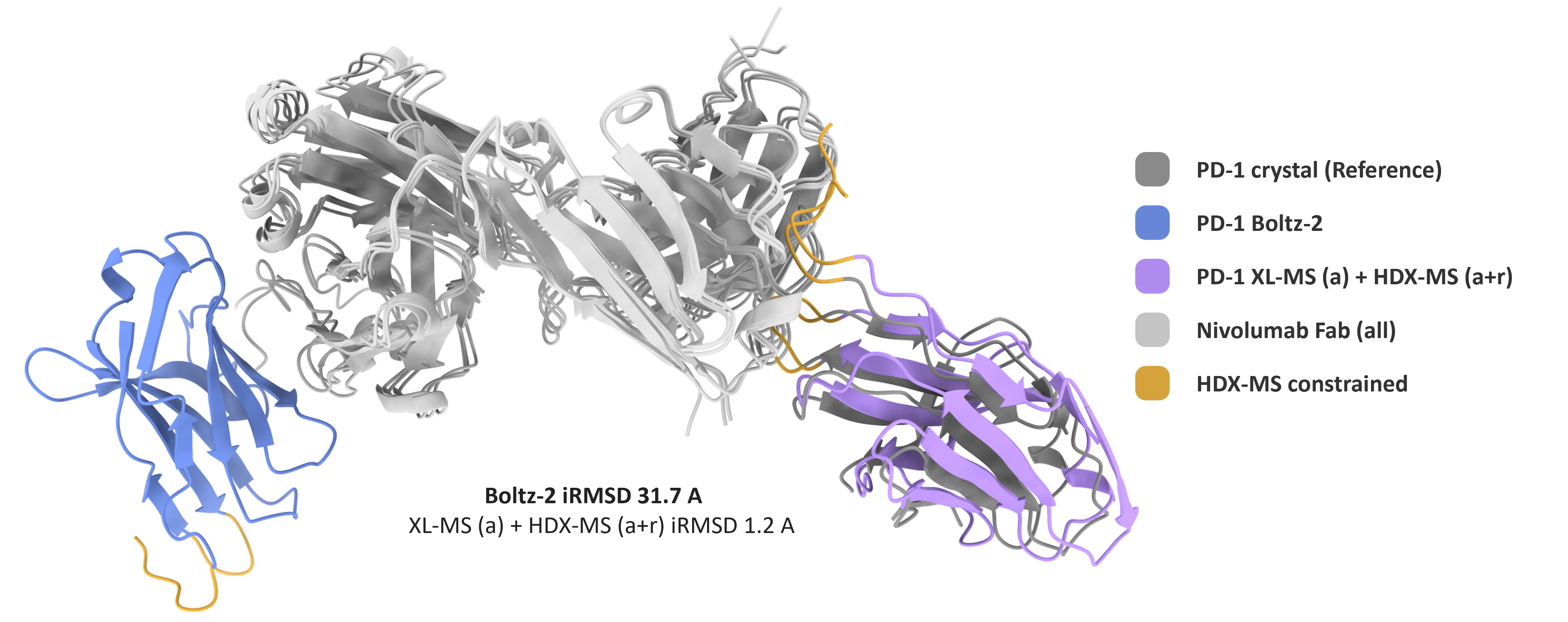}}
    \caption{PD-1-Nivolumab complex predictions. All structures are superposed on the Nivolumab Fab, shown in light grey for every model, so that differences reflect the placement of PD-1 relative to the antibody. PD-1 is shown in three copies - the crystal structure reference in dark grey, the unguided Boltz-2 prediction in blue, and the AIMS-Fold prediction in purple. Gold marks the HDX-MS constrained residues in each PD-1 structure.}
    \label{fig:pd1}
\end{figure}

Nivolumab binds PD-1 to block the deactivation of the host immune response, and the complex has both a solved crystal structure (5WT9) and published HDX-MS and XL-MS data \cite{Zhang2020}. Boltz-2 fails to predict the binding interface, positioning PD-1 on the wrong side of the antibody (Table~\ref{tab:aimsfold_benchmark_constraints}, Figure~\ref{fig:pd1}). HDX-MS constraints alone improve the DockQ and iRMSD scores to 0.23 and 4.4 \AA, and XL-MS alone to 0.69 and 1.3 \AA. Adding attractive HDX-MS constraints to the XL-MS, slightly reduces the score (0.64, 1.3 \AA), whereas including the repulsive HDX-MS constraints as well gives the best prediction for this system (0.71, 1.2 \AA). The average DockQ and interface RMSD values across the five generated models further support this (Table~\ref{tab:aimsfold_benchmark_constraints}).

KAT6A-CRBN is a novel system, with structural MS data we generated and no solved structure, so it is scored on constraint satisfaction alone. With the full constraint set AIMS-Fold raises the satisfied XL-MS constraints from 62 to 88\% and the satisfied HDX-MS constraints from 70 to 80\% (Table~\ref{tab:aimsfold_benchmark_constraints}). Its predictions were then used to prospectively rank PROTACs in a degrader campaign \cite{Arad2026}.

The same pattern holds across all systems (Table~\ref{tab:aimsfold_benchmark_constraints}). Unguided top-1 predictions reach DockQ 0.00--0.06, while the best guided condition reaches 0.43--0.71 across the four in-depth systems. The stronger modality varies between systems: PD-1-Nivolumab is better recovered by XL-MS, where HDX-MS alone reaches only 0.23, while BRD4-CRBN is recovered by HDX-MS, where attractive XL-MS alone reaches 0.21 (Table~\ref{tab:aimsfold_benchmark_constraints} and Figure~\ref{fig:brd4}). Combining all available constraint types yields the lowest iRMSD in three of the four systems.

Repulsive constraints alone recover the correct conformations on some systems. On BRD4-CRBN the repulsive cross-links clearly exceed the attractive ones (DockQ 0.35 versus 0.21), and the repulsive HDX-MS constraints match them (0.43 versus 0.42). On WDR5-DCAF1 the repulsive cross-links match the attractive ones (DockQ 0.50 versus 0.48). On PTPN2-CRBN and PD-1-Nivolumab the repulsive constraints alone do not improve on the unguided prediction.

Repulsive constraints show higher variance where they are effective. On WDR5-DCAF1, for example, the best of the five models generated with repulsive cross-links reaches DockQ 0.496, while the mean DockQ across the five generated models is 0.146 $\pm$ 0.182 (mean $\pm$ SD). This is expected, as excluding positions leaves a larger search space than specifying them. Taken together, these results show that structural-MS constraints allow the generation of more accurate conformations, and that the repulsive signal contributes to this in several systems.

\begin{figure}[ht]
    \centering
    \includegraphics[width=\linewidth]{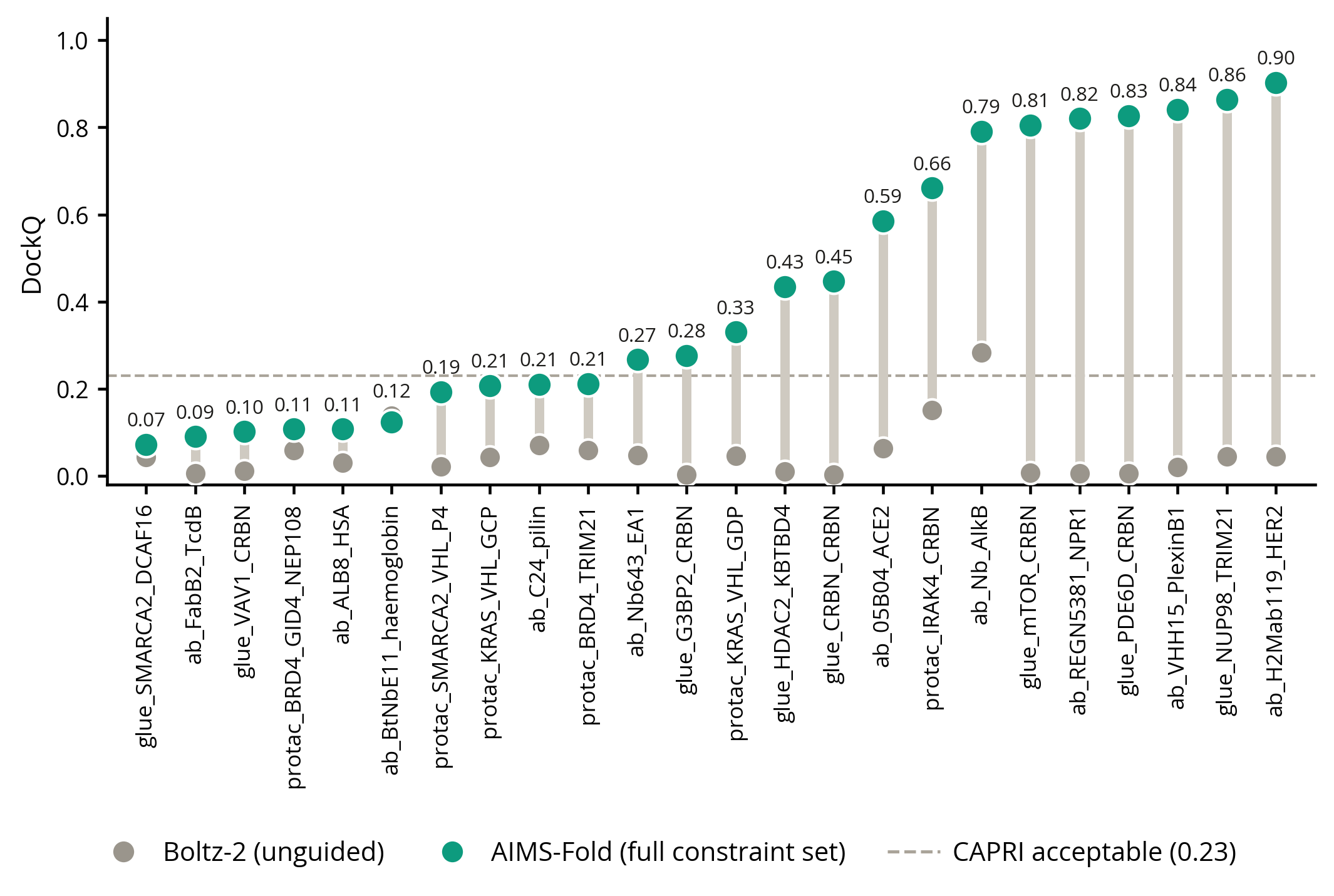}
    \caption{Test of AIMS-Fold on 24 further complexes where Boltz-2 fails. Shown is the DockQ of the best of five models. The five systems of Table~\ref{tab:aimsfold_benchmark_constraints}, where all constraint ablations were run, are not shown. Gray is unguided Boltz-2 and green is AIMS-Fold with the full constraint set (Methods). The bar joins the two values of a complex and the dashed line marks DockQ 0.23. Complexes are ordered by the AIMS-Fold result. The constraints are synthetic, derived from each deposited structure and the unguided model.}
    \label{fig:benchmark24}
\end{figure}

For the additional 24 complexes (10 antibody, 8 molecular glue and 6 PROTAC complexes), where Boltz-2 also fails, we used only the full constraint set, with synthetic constraints derived by comparing the deposited structure with the unguided Boltz-2 model. The number of complexes with an acceptable best model rises from 1 of 24 for unguided Boltz-2 to 14 of 24 for AIMS-Fold (Figure~\ref{fig:benchmark24}).

\begin{figure}[ht]
    \centering
    \includegraphics[width=\linewidth]{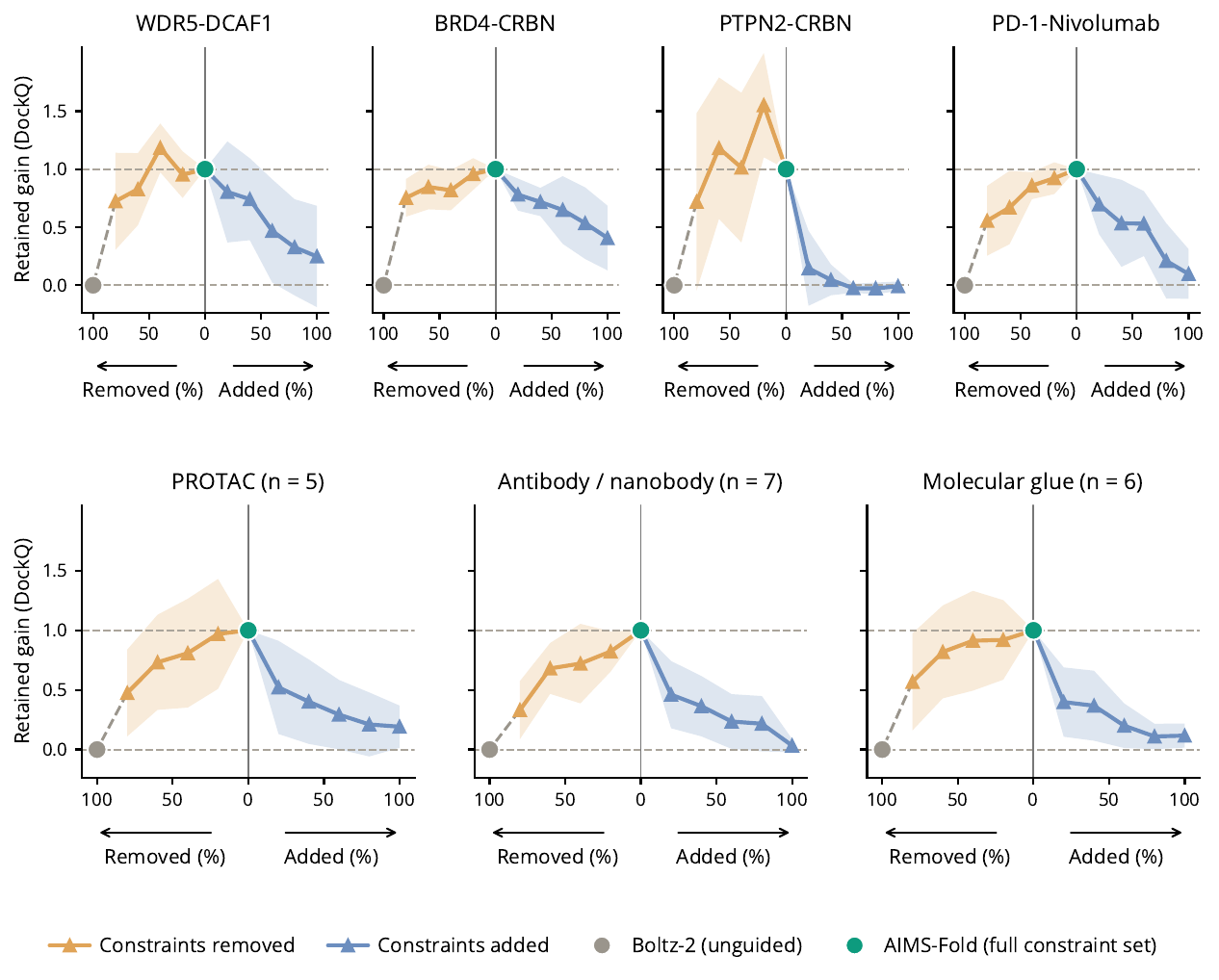}
    \caption{Retained AIMS-Fold gain under constraint dropout and noise. Top row, the four in-depth systems, where lines show the mean over ten random sets and bands one standard deviation across sets. Bottom row, the 18 of 28 systems in which AIMS-Fold reaches a DockQ above 0.23, by modality (PROTAC 5, antibody or nanobody 7, glue 6). Here each system's value is the mean over the ten sets, and lines and bands show the mean and the standard deviation across systems. Retained gain is the DockQ gain over Boltz-2 at each perturbation level divided by the gain of the unperturbed AIMS-Fold run (1 is unperturbed, 0 is Boltz-2 without constraints). Corruption was applied to the full attractive and repulsive constraint set. Left of 0, a random fraction of the constraints was withheld. Right of 0, random decoy constraints were added (random lysine pairs and random solvent-exposed residues). At each level, ten random constraint sets were drawn. The gray point is unguided Boltz-2 and the green point the unperturbed AIMS-Fold run.}
    \label{fig:noise}\label{fig:retained_gain}
\end{figure}

To assess how robust AIMS-Fold is to incomplete or imperfect constraint sets, we perturbed the constraints in two ways, dropout, which withholds a random fraction of them, and noise, which adds the same fraction of random decoy constraints. We varied the fraction from 20 to 100\%, drew ten random constraint sets at each level and compared DockQ with the unperturbed run. The mean DockQ stays acceptable (>0.23) at every dropout level on all four in-depth systems (Figure~\ref{fig:noise}). With added noise it stays acceptable at every level on WDR5-DCAF1 and BRD4-CRBN and up to 60\% on PD-1-Nivolumab, while PTPN2-CRBN falls below 0.23 from 40\%. On the 18 of 28 complexes where AIMS-Fold reaches an acceptable DockQ, removing constraints reduces it gradually, while adding random constraints erodes it faster (Figure~\ref{fig:retained_gain}). Thus, while the model is more sensitive to false positives than to missing data, it maintains a substantial degree of robustness against both. Guidance could in principle distort the structures, so we also compared clashes and backbone angle deviations of guided and unguided models and found them to be comparable (Supplementary Figure~\ref{fig:geometry}).

\subsection{Guided generation provides an advantage over post-hoc filtering}

Historically, the default strategy for integrating experimental data into computational modeling was to use structural constraints to score and filter a large ensemble of naive predictions post-generation. This relies on the assumption that the unconstrained model sampled the correct conformation in the first place. In vast and flexible conformational spaces such as PROTAC ternary structures, unconstrained models can collapse into static states memorized during training, and if the correct basin is never explored, filtering thousands of predictions yields nothing. Guided generation removes that dependency by altering the probability landscape during generation rather than selecting after it.

\begin{figure}[ht]
    \centering
    \includegraphics[width=\linewidth]{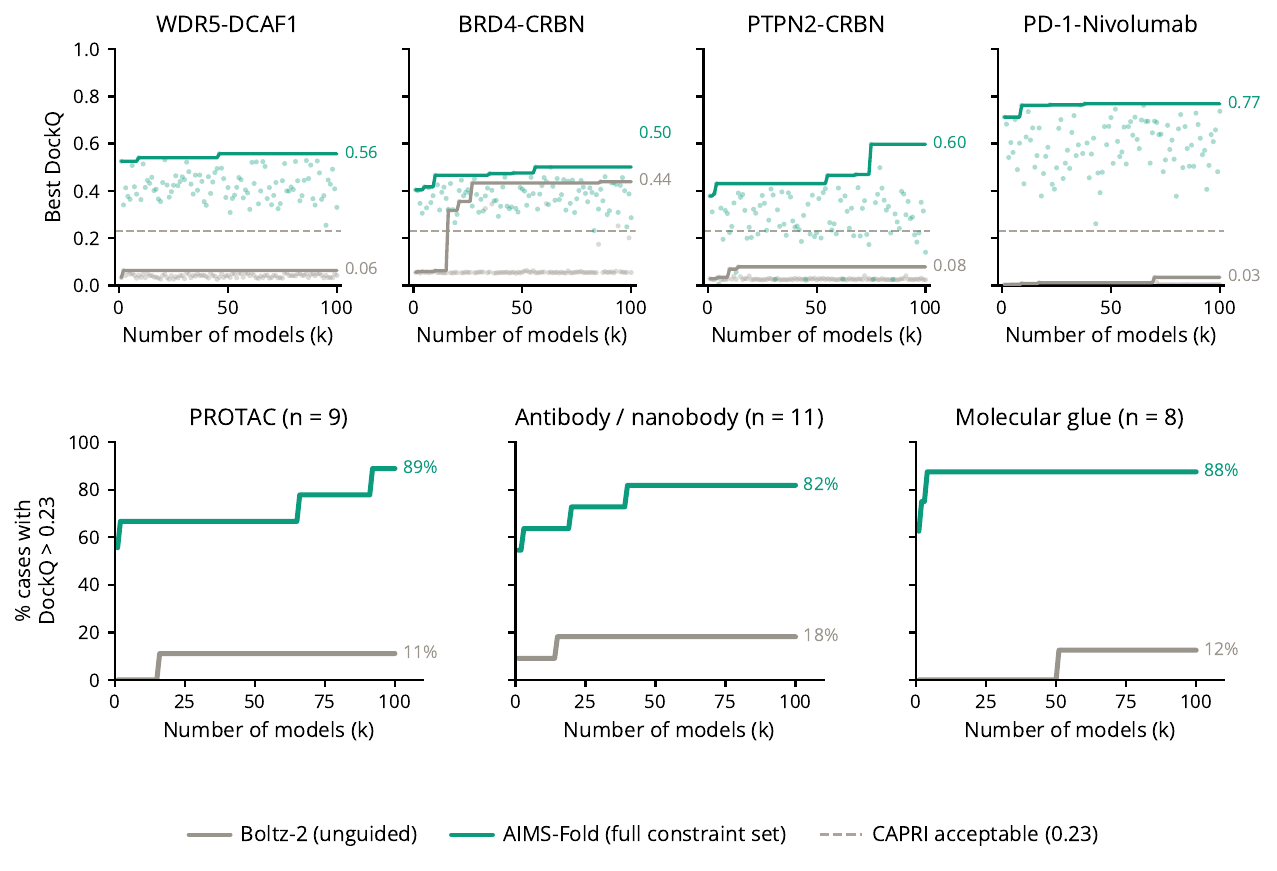}
    \caption{Constraints, not sampling, drive the gain in prediction accuracy across all three modalities. Unguided Boltz-2 is gray and AIMS-Fold with the full constraint set green. Each complex was predicted with 20 seeds and 5 diffusion samples per seed, 100 models per complex per method, and best-of-k is the running maximum DockQ over models in a fixed draw order. Top row, the four in-depth systems: points are individual models, the step lines the best DockQ among the first k models, and the number at the end of each line the best over all 100. The dashed line is DockQ 0.23. Bottom row, the fraction of the 28 complexes (four in-depth systems and the 24 additional cases) reaching a CAPRI acceptable score (DockQ 0.23) against the number of models drawn, split by modality. At 100 models AIMS-Fold reaches a CAPRI acceptable score on 82 to 89\% of the complexes in each modality, against 11 to 18\% for unguided Boltz-2.}
    \label{fig:sampling28}
\end{figure}

To test whether this applies to our cases, we generated 100 models with Boltz-2 and AIMS-Fold for each of the 28 complexes with a reference structure, across 20 seeds, and tracked the best model as a function of the number generated (Figures~\ref{fig:sampling28}, \ref{fig:sample_size_dockq}). More sampling does not close the gap: at 100 models unguided Boltz-2 reaches an acceptable conformation for 11 to 18\% of complexes per modality, against 82 to 89\% for AIMS-Fold. This stark contrast shows that for non-canonical complexes, inference-time guidance is not just a refinement, it is fundamentally necessary for finding the correct structure.

\section{Discussion}

We introduce AIMS-Fold, an inference-time guidance layer that steers a pretrained co-folding model with constraints derived from XL-MS and HDX-MS. Across 28 complexes with a reference structure, guidance often recovers an acceptable conformation that the unguided model does not reach, and a single guided model reaches one more often than 100 unguided ones (Figure~\ref{fig:sampling28}).

Structure prediction of PROTAC-based ternary complexes is a challenge for existing models, and XL-MS and HDX-MS are well suited to address this problem. XL-MS constrains how the proteins are oriented relative to one another and HDX-MS marks which surfaces are involved. Together, these inform whether the target lysines end up close enough to the ligase machinery for ubiquitination to occur.

Across the benchmark, high DockQ values ($>$0.8) occur only for antibody and glue complexes, and no PROTAC reaches this range (Figure~\ref{fig:benchmark24}), reflecting the smaller, linker-bridged interfaces. The fraction of native contacts DockQ depends on is noisy when few contacts exist, so the score understates a correct orientation, which the iRMSD confirms in the in-depth systems (Table~\ref{tab:aimsfold_benchmark_constraints}). Since ubiquitination depends on orientation, predictions at this level of accuracy suffice to assess whether it can occur. The KAT6A-CRBN predictions, generated using AIMS-Fold for a system with no solved structure, were used prospectively to rank PROTACs in a degrader optimization campaign that produced an \textit{in vivo} validated compound \cite{Arad2026}.

Repulsive cross-links come from a signal that current co-folding models cannot take as input. When cross-linking is run across compounds, pairs whose intensity drops with one PROTAC relative to another indicate residues that are further apart in that state. Existing co-folding models accept only attractive signals, leaving this information unused. Remarkably, in the BRD4-CRBN case, the repulsive cross-links alone outperform the attractive ones.

The same holds for HDX-based constraints. Peptides with good coverage and no uptake difference between states indicate regions that remain exposed in the complex, and this too cannot be supplied to current co-folding models. Our results suggest these regions carry a usable signal: on the synthetic cases the repulsive constraints improve the prediction on their own, and on the PD-1-Nivolumab case they improve it when combined with the XL-MS data.

Repulsive and attractive constraints are not equivalent. An attractive constraint states where a residue belongs, which narrows the search considerably. A repulsive constraint only rules positions out, leaving open a larger search space. This is consistent with what we observe: where repulsive constraints improve the prediction they often do so with markedly higher variance, reaching the correct conformation occasionally rather than consistently (Table~\ref{tab:aimsfold_benchmark_constraints}). It also makes constraint satisfaction a weaker criterion for ranking generated models, since a requirement that two residues stay apart can be met by a vast range of structures. For example, on PD-1-Nivolumab the unguided predictions already satisfy all repulsive constraints while remaining entirely wrong. Finally, adding repulsive constraints on top of attractive ones does not consistently improve the prediction.

Constraints derived from a deposited structure are cleaner than measured ones, so the 28-complex benchmark gives an upper bound on what guidance can do. The noise and dropout tests, on the four in-depth systems and on the 18 complexes where AIMS-Fold reaches an acceptable DockQ, show that while the model tolerates missing data well, it degrades more noticeably under false positive noise. Measured data can therefore be used even when coverage is partial, as long as the constraints that are used are mostly reliable. Still, more systems with measured structural MS data are needed to verify the model further.

Finally, XL-MS and HDX-MS measure an ensemble of conformations, which fits well in the case of PROTAC ternary complexes, which are highly dynamic \cite{Rovers2024}. Here we use the data to select one state, but the same signal should support the generation of an ensemble, with subsets of mutually consistent constraints producing distinct conformations. We intend to pursue this in future work.

\bibliographystyle{unsrtnat}
\bibliography{references}

\appendix
\renewcommand{\figscale}{1}
\setcounter{figure}{0}
\renewcommand{\thefigure}{S\arabic{figure}}
\setcounter{table}{0}
\renewcommand{\thetable}{S\arabic{table}}

\section{Methods}

\subsection{MS-guided diffusion and steering}

AIMS-Fold guides a pretrained co-folding model, Boltz-2, with structural proteomics constraints at inference time and requires no retraining. Each experimental constraint is translated into a differentiable potential whose gradient is injected into the sampling trajectory during denoising, reshaping the sampling distribution.

\subsection{XL-MS constraints}

XL-MS data is incorporated as distance constraints. Cross-links yield two types of spatial constraints. Positive constraints indicate that two residues are within the reach of the cross-linker under a given experimental condition; these are already implemented in Boltz-2. Negative constraints assume that the distance between two residues exceeds that reach. These negative constraints arise from differential XL-MS studies where a cross-link is observed in one state and significantly depleted in another, indicating that the cross-linker cannot bridge the two residues in the second state.

We implement negative constraints as a negation flag on Boltz-2's existing flat-bottom distance potential, which inverts its bound from an upper to a lower one. For a set of negatively constrained residue pairs $\mathcal{N}_{\text{neg}}$, the repulsive potential is formulated as:
\begin{equation*}
U_{\mathrm{neg}}(\mathbf{x}) = k \sum_{(i,j) \in \mathcal{N}_{\mathrm{neg}}} \max\!\left(0,\; d_{\min} - d_{ij}\right)
\end{equation*}
where $d_{ij}$ is the distance between residues $i$ and $j$, and $d_{\min}$ is the reach of the cross-linker. The potential and its gradient are zero once the pair is separated. An unsatisfied pair is pushed apart at constant force throughout denoising. Each pair is penalised independently, so all negative constraints must be satisfied simultaneously.

\subsection{HDX-MS constraints}

HDX-MS data is incorporated as protection constraints. Relative fractional uptake reports the change in deuterium exchange between bound and unbound states (in the case of ternary structures it is ternary vs binary or apo), where negative values indicate protection upon complex formation and positive values indicate residues that remain exposed. In contrast to cross-linking measurements, HDX-MS data is unpaired, reporting that a given residue was shielded from solvent without specifying the interacting entity responsible for the shielding. Protection therefore cannot be expressed as a residue-pair distance, and is instead mapped onto the geometry between a residue and a set of candidate partner chains, which Boltz-2 expresses as a pocket constraint.

Boltz-2 provides a pocket constraint, which requires a set of residues on one chain to lie within a given distance of a single named binder chain. This is close to what protection data describes, but it presumes the covering chain is known. We therefore extended the pocket constraint to accept a set of candidate partner chains, aggregated disjunctively, so that the constraint is satisfied when \emph{any} one of them covers the residue. Each protected residue generates one such constraint. For a set of protected residues $\mathcal{P}$, the attractive potential is
\begin{equation*}
U_{\mathrm{prot}}(\mathbf{x}) = k \sum_{i \in \mathcal{P}}
  \max\!\left(0,\; d_i - d_{\max}\right),
  \qquad d_i = \min_{a \in \mathcal{C}(i)} d_{ia},
\end{equation*}
where $\mathcal{C}(i)$ is the set of atoms in the partner chains designated for residue $i$, and $d_{ia}$ is the distance between residue $i$ and atom $a$. Boltz-2 aggregates a contact constraint over the atoms of one partner chain while we extend this aggregation across several candidate chains, so that a single constraint is satisfied when \emph{any} one of them is in contact. The aggregation is a smooth minimum that sharpens over the trajectory, so the disjunction is diffuse while the structure is still noisy and becomes decisive as it resolves. In a ternary complex the protecting entity may be the second protein or the ligand, and the measurement does not say which, so the constraint must admit either.

We use the negation flag introduced for XL-MS to express the opposite signal. Residues that remain exposed invert the bound, and for a set of deprotected residues $\mathcal{D}$,
\begin{equation*}
U_{\mathrm{dep}}(\mathbf{x}) = k \sum_{i \in \mathcal{D}} \sum_{a \in \mathcal{C}(i)}
  \max\!\left(0,\; d^{\,\mathrm{dep}}_{\min} - d_{ia}\right).
\end{equation*}
Here the aggregation is conjunctive rather than disjunctive. A protected residue needs contact with only one partner to satisfy its constraint, whereas an exposed surface must stay clear of all of them, so every partner atom is penalised independently.

\subsection{Setup and configuration}

Protection and deprotection constraints each use a single distance bound, applied identically in every system: $d_{\max} = 3.0$~\AA{} and $d^{\,\mathrm{dep}}_{\min} = 8.8$~\AA. All other settings are Boltz-2 defaults, shared by both modalities and unchanged across systems: evaluation interval 4, guidance weights 0 / 0.5 / 1.0 across timestep bands, union $\lambda$ interpolating $8.0 \rightarrow 0.0$ at $\alpha = -2.0$.

\subsection{Per case constraints selection}

BRD4-CRBN and WDR5-DCAF1 (synthetic). These two cases have two conformations with solved structures. Cross-links were derived based on the difference between the two. For every lysine-lysine pair we measured the C$\alpha$-C$\alpha$ distance in both structures against the 30 \AA{} reach of a conventional cross-linker, discarding pairs with the same outcome in both, and among the rest keeping only those whose distance changes by more than 10 \AA. Surviving pairs became positive constraints when the residues were within reach (under 30 \AA) in the target conformation, and became repulsive constraints when they were within reach in the other resolved structure but beyond reach in the target conformation. HDX-MS constraints were selected based on the same principle, keeping residues with differential $\Delta$RSA between the two structures.

PTPN2-CRBN (experimental, HDX-MS only; no XL-MS data available). HDX-MS data was collected only for PTPN2 \cite{Hao2024}. The four residues with the strongest ternary-minus-apo protection, using a $>$5\% protection threshold were selected as positive constraints. Negative HDX-MS constraints were selected as surface exposed residues with $<$5\% protection in ternary-apo, and within a radius of 8 \AA{} from the PTPN2 binding pocket.

PD-1-Nivolumab (experimental, both modalities, both previously published \cite{Zhang2020}). A subset of solvent-exposed residues from the two most significantly decreasing peptides were chosen as positive HDX-MS constraints, while all interface residues in the expected epitope and paratope of the binding with $<$5\% protection were chosen as negative constraints. For XL-MS constraints we used all the cross-links reported in the same study.

KAT6A-CRBN (experimental, both modalities). This complex has no solved structure, so both constraint sets come from in-house data acquired with an active and an inactive compound, and both were derived by the same differential logic. Cross-links were assigned as positive where intensity is significantly enriched in the active-compound condition and negative where it is significantly depleted relative to the inactive one. HDX-MS constraints were taken as the residues showing differential ternary-minus-apo protection between the same two samples, so that in both modalities the constraint encodes what changes between a productive and an unproductive ternary complex rather than absolute geometry.

\subsection{Constraint satisfaction}

Satisfaction is a binary per-residue test on $\Delta$RSA in the predicted complex with a threshold of 0.05 RSA units. A negative hdx-constrained (deprotected) residue counts as satisfied when $\Delta$RSA $<$ 0.05, meaning it buries almost no surface on binding, and a positive (protected) hdx-constrained residue is satisfied when $\Delta$RSA $>$ 0.05. Similarly, XL-MS constraints work on C$\alpha$-C$\alpha$ distance with a 30 \AA{} threshold, satisfied below it for attract pairs and above it for repel pairs. We report the fraction satisfied over the union of all constraints for that target, for the top-1 model and as mean $\pm$ SD across the five models. Models were ranked by the highest percentage of fulfilled constraints, ties broken by DockQ; where no reference structure was available (KAT6A) we took the first generated model among the tied candidates.

\subsection{DockQ, iRMSD, lRMSD.}

We report DockQ \cite{Basu2016} alongside interface and ligand RMSD (iRMSD and lRMSD, respectively), and treat DockQ $\geq$ 0.23 as acceptable, following the CAPRI criteria used in previous PROTAC benchmarks \cite{Pereira2025-va}. We used the official DockQ implementation (\url{https://github.com/wallnerlab/DockQ}). For antibody complexes we changed the aggregation to include only antibody-antigen interfaces.

\subsection{Local geometry}

Local geometry was assessed by heavy-atom clash counts (overlap above 0.4 \AA{}, 2.5 \AA{} cut-off for N/O pairs) and backbone bond-length and bond-angle deviations from Engh and Huber ideal values, on all 100 models per condition. Each model was scored only over residues resolved in the reference structure, found by sequence alignment of each model chain to its reference chain, so both conditions are compared over the same residues. A chain whose reference is a different construct (BRD4-CRBN) was scored in full.

\subsection{Additional benchmark cases}

Of the 28 complexes with a reference structure, four were studied in depth. The other 24 are recent PDB entries of PROTAC ternary, molecular glue and antibody or nanobody complexes for which unguided Boltz-2 fails to reproduce the interface. Twenty were released after the Boltz-2 training cutoff. The other four, like PD-1-Nivolumab, predate it and Boltz-2 still fails on them. We used only the full constraint set for these 24. The constraints are synthetic. They were derived by comparing the deposited structure with the unguided model, so each constraint is met by the deposit and violated by the unguided pose. HDX-MS attract and repel residues come from the change in relative solvent accessibility, at most 20 per case. XL-MS attract and repel pairs come from the C$\alpha$-C$\alpha$ distance against the 30~\AA{} bound, at most 10 per case. For the sampling analysis, each complex was predicted with 20 seeds and 5 diffusion samples per seed, giving 100 models per complex per method, and best-of-k was taken as the running maximum DockQ over models in a fixed seed order. Complexes were assigned to one of three modalities, PROTAC ternary complex, molecular glue complex, or antibody or nanobody complex, from the composition of the deposit.

\section{Supplementary figures}
\begin{figure}[ht]
    \centering
    \includegraphics[width=\linewidth]{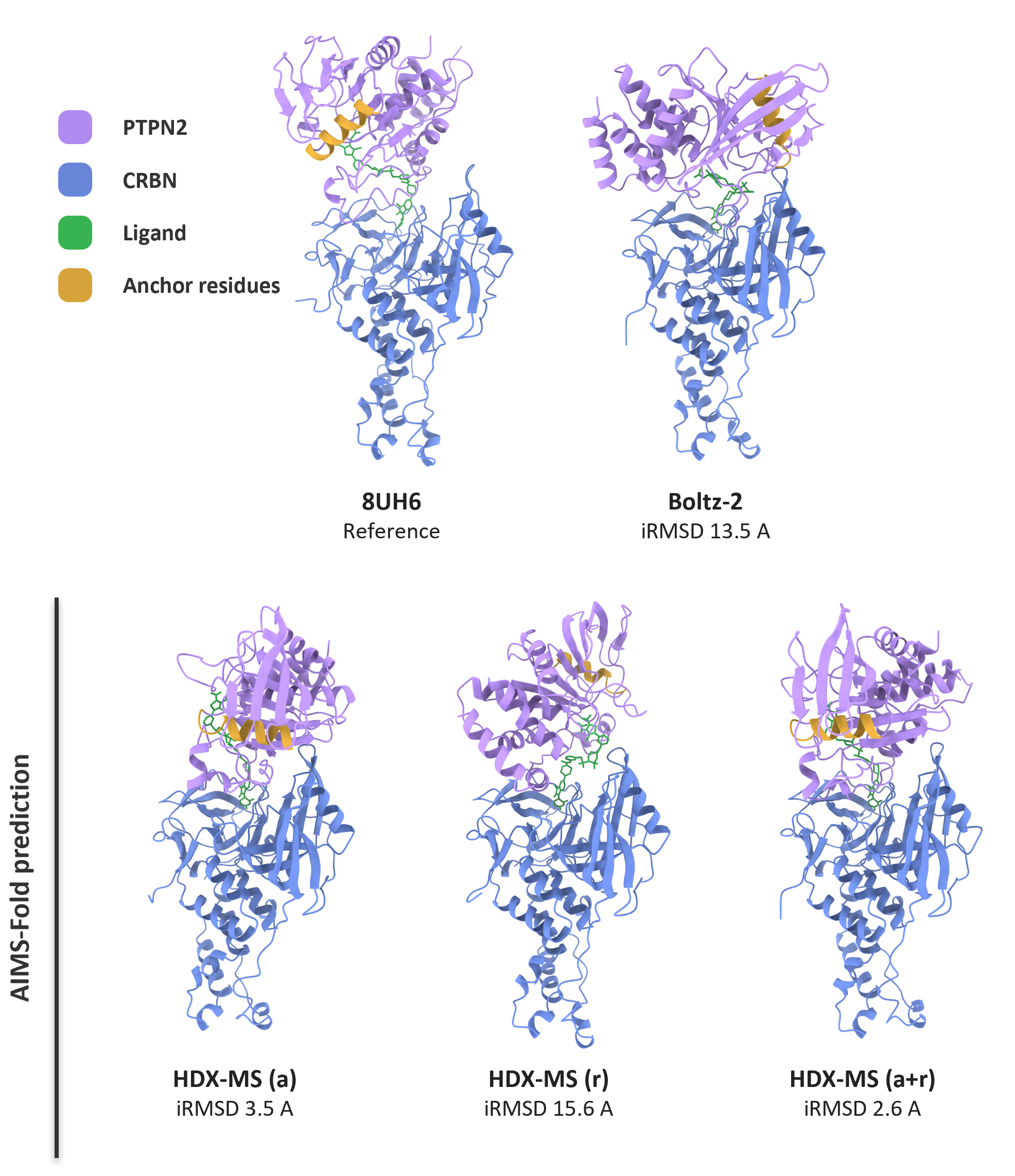}
    \caption{PTPN2-PROTAC-CRBN ternary complex predictions. All structures are superposed on CRBN and shown in a common orientation. PTPN2 is coloured purple, CRBN blue, and the PROTAC green. Gold stretches mark a reference segment on each chain, included for easy comparison across panels. The top row shows the cryo-EM structure 8UH6, the reference against which all predictions are scored, and the unguided Boltz-2 prediction. The panels below are AIMS-Fold predictions guided by HDX-MS constraints applied as attractive only (a), repulsive only (r), or both together (a+r); no XL-MS data were available for this case. iRMSD to 8UH6 is given below each model.}
    \label{fig:ptpn2}
\end{figure}
\begin{figure}[ht]
    \centering
    \includegraphics[width=\linewidth]{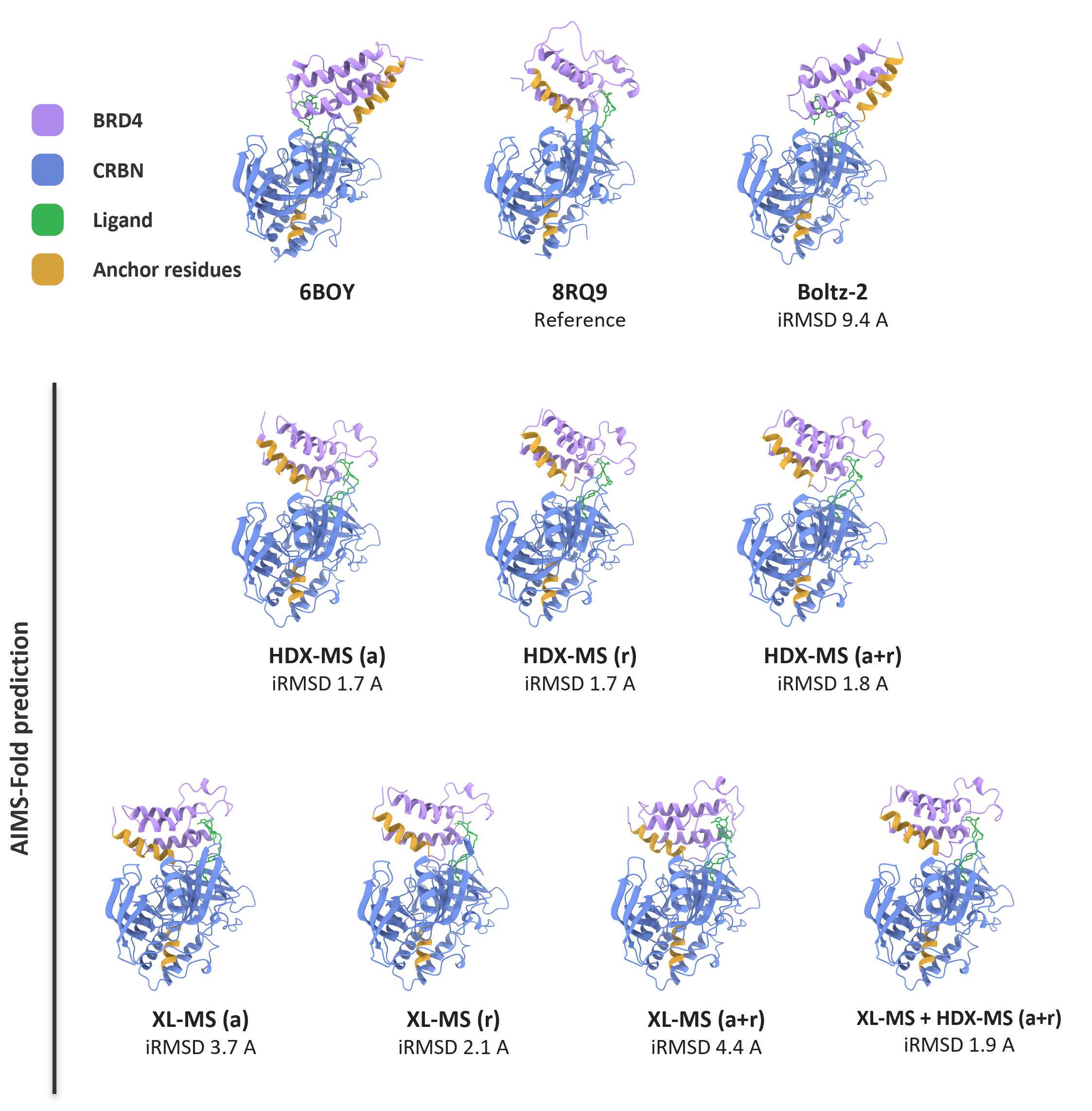}
    \caption{BRD4(BD1)-PROTAC-CRBN ternary complex predictions. All structures are superposed on CRBN and shown in a common orientation. BRD4 is coloured purple, CRBN blue, and the PROTAC green. Gold stretches mark a reference segment on each chain, included for easy comparison across panels. The top row shows two crystal structures of the complex, 6BOY and 8RQ9, followed by the unguided Boltz-2 prediction. 8RQ9 is the reference against which all predictions are scored. The panels below are AIMS-Fold predictions under different constraint sets - HDX-MS and XL-MS, each applied as attractive only (a), repulsive only (r), or both together (a+r), and the two modalities combined, XL-MS + HDX-MS (a+r). iRMSD to 8RQ9 is given below each model.}
    \label{fig:brd4}
\end{figure}
\begin{figure}[htbp]
    \centering
    \textbf{A}\par\smallskip
    \includegraphics[width=\linewidth]{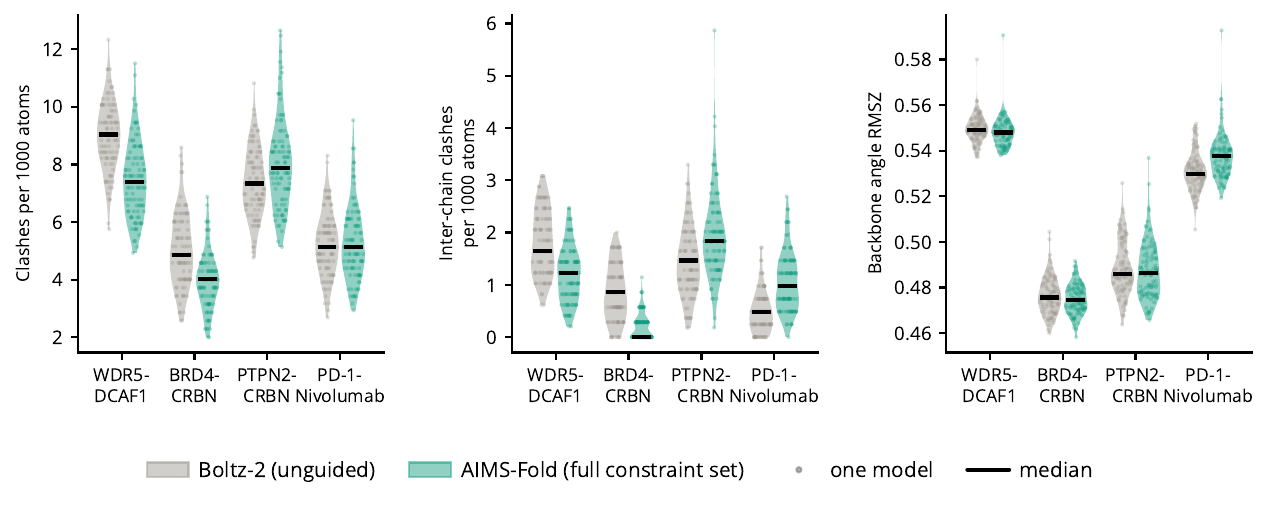}\par\bigskip
    \textbf{B}\par\smallskip
    \includegraphics[width=\linewidth]{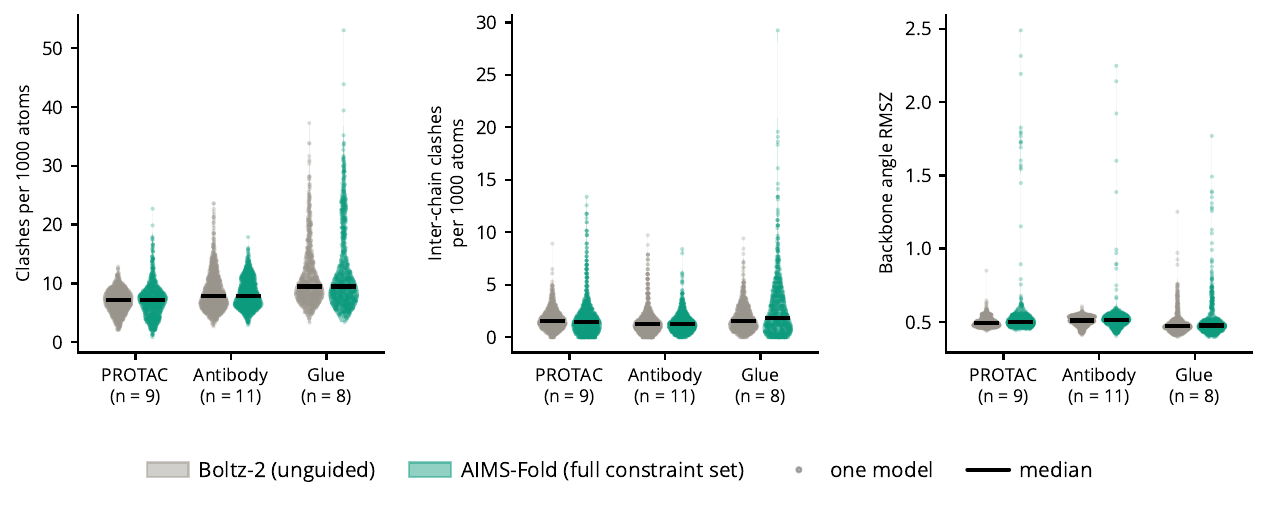}
    \caption{Local geometry of unguided and guided models for (A) the four in-depth systems, 100 models per condition, and (B) all 28 systems by modality (PROTAC n = 9, antibody n = 11, glue n = 8), pooling all 100 models of each system. Gray is Boltz-2 (unguided) and green AIMS-Fold (full constraint set). Every dot is one model and the black bar is the median. Left, heavy-atom clashes per 1000 atoms. Middle, the same count restricted to clashes between chains. Right, the root-mean-square z-score (RMSZ) of backbone bond angles against Engh and Huber ideals. Each model was scored only over residues resolved in its reference structure.}
    \label{fig:geometry}
\end{figure}
\begin{figure}[p]
    \centering
    \includegraphics[width=\linewidth,height=0.78\textheight,keepaspectratio]{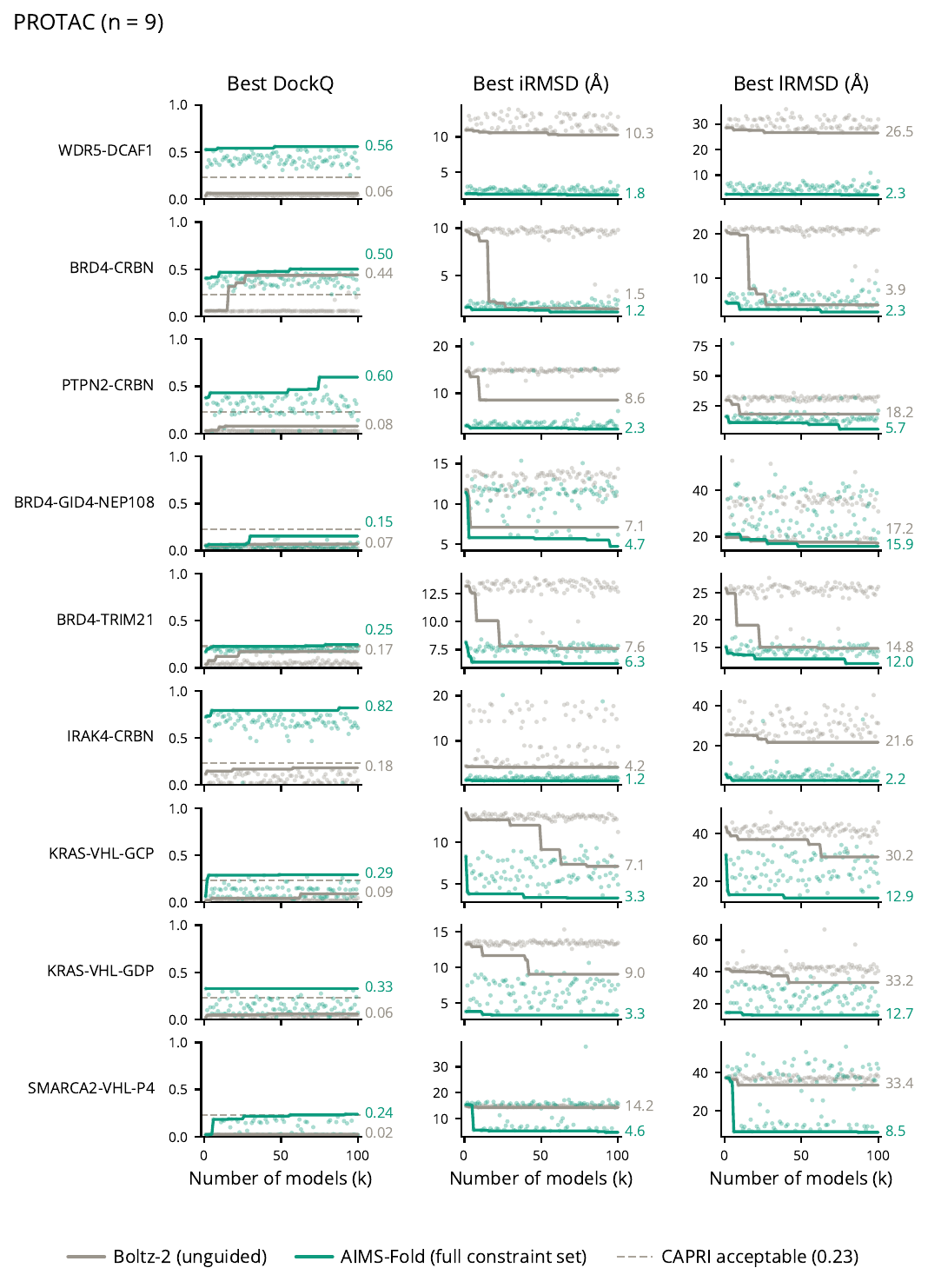}
    \caption{Best model as a function of sample size for all 28 complexes, one page per modality: (A) PROTAC, (B) antibody or nanobody, (C) molecular glue. Each row is one complex. Columns show the best DockQ, iRMSD and lRMSD as a function of the number of generated structures of unguided Boltz-2 (gray) and AIMS-Fold (green) using the full set of available constraints. Points indicate the scores of individual models and the solid step lines the running best, pooled over 20 independent seeds giving 100 models in total. The best score obtained over all 100 models is annotated at the end of each line. The dashed line is DockQ 0.23. (A) PROTAC complexes.}
    \label{fig:sample_size_dockq}\label{fig:rmsd_curves}
\end{figure}
\begin{figure}[p]
    \addtocounter{figure}{-1}
    \centering
    \includegraphics[width=\linewidth,height=0.85\textheight,keepaspectratio]{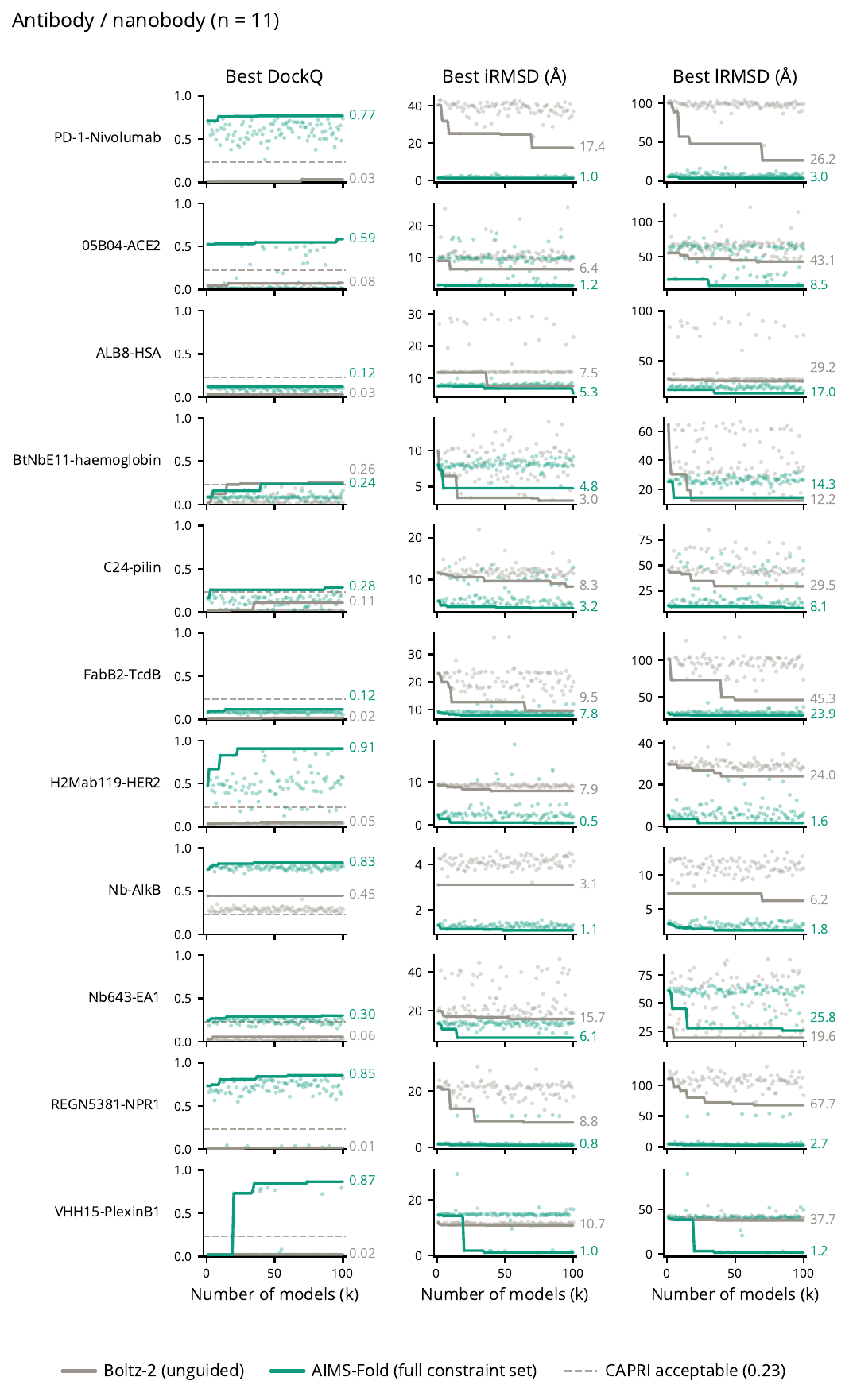}
    \caption{(continued) (B) Antibody and nanobody complexes.}
\end{figure}
\begin{figure}[p]
    \addtocounter{figure}{-1}
    \centering
    \includegraphics[width=\linewidth,height=0.85\textheight,keepaspectratio]{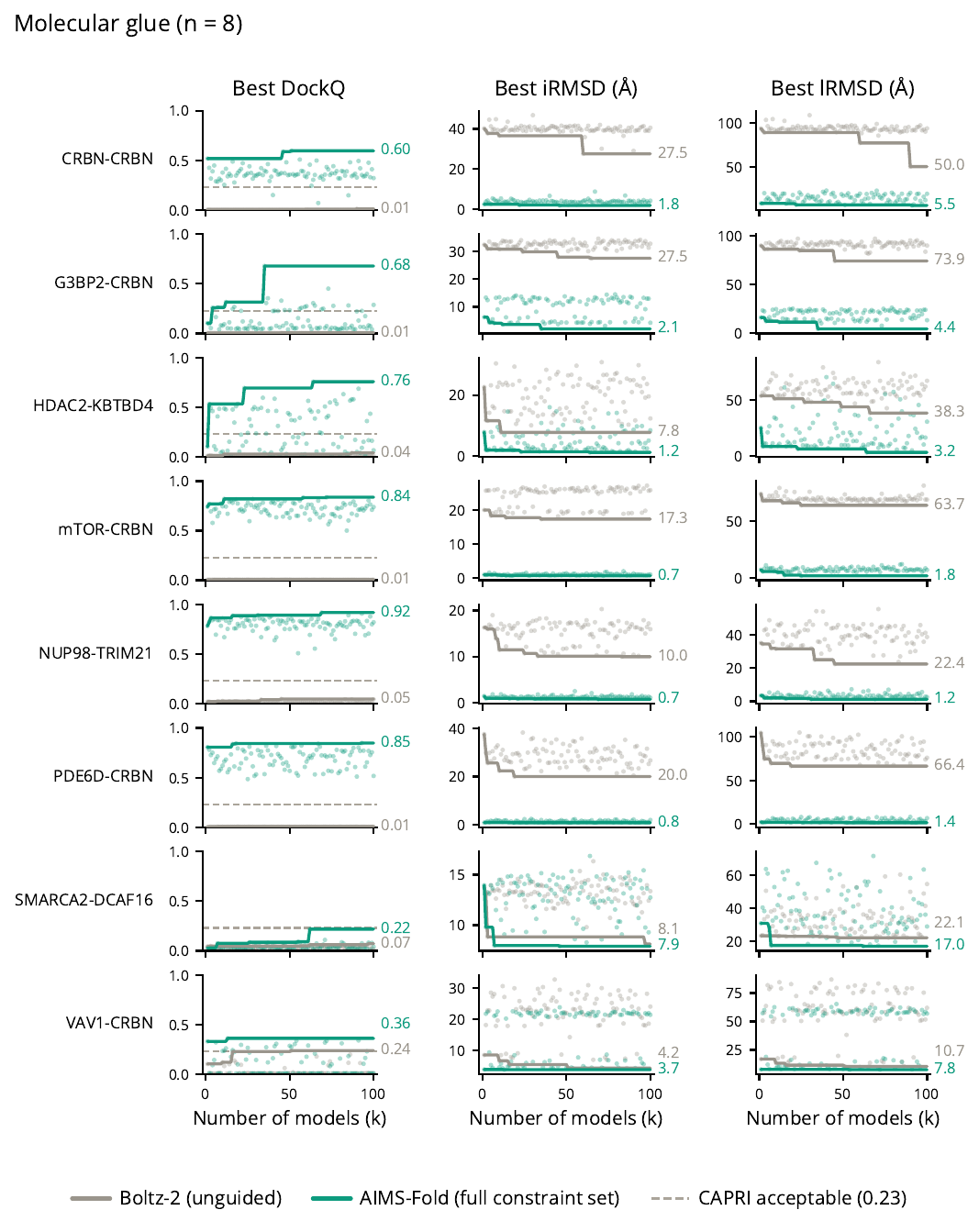}
    \caption{(continued) (C) Molecular glue complexes.}
\end{figure}

\end{document}